\documentclass[conference]{IEEEtran}

\makeatletter

\def\ps@IEEEtitlepagestyle{%
  \def\@oddfoot{\mycopyrightnotice}%
  \def\@evenfoot{}%
}
\def\mycopyrightnotice{%
  {\footnotesize 979-8-3503-9452-8/24/\$31.00~\copyright~2024 European Union\hfill}
  \gdef\mycopyrightnotice{}
}

\usepackage{blindtext}
\usepackage{eso-pic}
\IEEEoverridecommandlockouts
\usepackage{amsmath,amssymb,amsfonts}

\usepackage{graphicx}
\usepackage{textcomp}
\usepackage{xcolor}
\def\BibTeX{{\rm B\kern-.05em{\sc i\kern-.025em b}\kern-.08em
    T\kern-.1667em\lower.7ex\hbox{E}\kern-.125emX}}
    
\usepackage{eso-pic}
\newcommand\AtPageUpperMyright[1]{\AtPageUpperLeft{%
 \put(\LenToUnit{0.17\paperwidth},\LenToUnit{-2cm}){%
     \parbox{0.9\textwidth}{\raggedleft\fontsize{8}{11}\selectfont #1}}%
 }}%
\newcommand{\conf}[1]{%
\AddToShipoutPictureBG*{%
\AtPageUpperMyright{#1}
}
}    
\usepackage[utf8]{inputenc} 
\usepackage{hyperref}       
\usepackage{url}            
\usepackage{booktabs}       
\usepackage{nicefrac}       
\usepackage{microtype}      
\usepackage{xcolor}         
\usepackage{bm}
\usepackage[sorting=none,style=numeric,giveninits=true,doi=false,isbn=false,url=false,eprint=false]{biblatex}
\renewbibmacro*{issue+date}{%
	\setunit{\addcomma\space}
		\iffieldundef{issue}
		{\usebibmacro{date}}
		{\printfield{issue}%
			\setunit*{\addspace}%
	\usebibmacro{date}}
\newunit}
\bibliography{biblio}
\AtBeginBibliography{\footnotesize}

\usepackage{algorithm}
\usepackage{algpseudocode}
\usepackage{float}

\usepackage[font=footnotesize]{caption}
\usepackage{subcaption}
\usepackage[symbol]{footmisc}

\PassOptionsToPackage{printonlyused,smaller}{acronym}
\usepackage[nolist]{acronym} 

\newcommand{\ie}{i.\,e.\;}

\newcommand{\detv}[1]{\mathbf{#1}}

\newcommand{\detm}[1]{\mathbf{#1}}

\newcommand\txtw{0.98}

\newcommand{\change}[1]{#1} 

\newcommand\github{https://github.com/Probabilistic-ML/pirl}

\begin{document}
	\begin{acronym}
		\acro{BNN}{{B}ayesian neural network}
		\acro{CMC}{continuous mountain car}
		\acro{DGCNTS}{deep {G}aussian covariance network with trajectory sampling}
		\acro{DGCN}{deep {G}aussian covariance network}
		\acro{EKF}{extended {K}alman filter}
		\acro{GP}{{G}aussian process}
		\acrodefplural{GP}{{G}aussian processes}
		\acro{IDP}{inverted double pendulum}
		\acro{IPSU}{inverted pendulum swing up}
		\acro{MBRL}{model-based reinforcement learning}
		\acro{MPC}{model predictive control}
		\acro{NN}{neural network}
		\acro{PETS}{probabilistic ensembles with trajectory sampling}
		\acro{PILCO}{probabilistic inference for learning control}
		\acro{PIRL}{probabilistic inference for reinforcement learning}
		\acro{PNN}{probabilistic neural network}
		\acro{PPO}{proximal policy optimization}
		\acro{P}{pendulum}
		\acro{RBF}{radial basis function}
		\acro{UKF}{unscented {K}alman filter}
	\end{acronym}

\title{\vspace*{1cm} Deep Gaussian Covariance Network with Trajectory Sampling for Data-Efficient Policy Search
\thanks{Thanks to German Federal Ministry of Education and Research (BMBF) - Funding number: 13FH174PX8}
}

\author{\IEEEauthorblockN{Can Bogoclu\textsuperscript{*, $\dagger$}}
\IEEEauthorblockA{\textit{Zalando SE} \\
Berlin, Germany \\
ORCID: 0000-0002-4067-9949}
\and
\IEEEauthorblockN{Robert Vosshall\textsuperscript{*}}
\IEEEauthorblockA{\textit{Institute for Modelling and High Performance Computing} \\
\textit{Niederrhein University of Applied Sciences}\\
Krefeld, Germany \\
robert.vosshall@gmail.com}
\and
\IEEEauthorblockN{Kevin Cremanns}
\IEEEauthorblockA{\textit{Institute for Modelling and High Performance Computing} \\
\textit{Niederrhein University of Applied Sciences}\\
Krefeld, Germany \\
kevin.cremanns@hs-niederrhein.de}
\and
\IEEEauthorblockN{Dirk Roos}
\IEEEauthorblockA{\textit{Institute for Modelling and High Performance Computing} \\
\textit{Niederrhein University of Applied Sciences}\\
Krefeld, Germany \\
dirk.roos@hs-niederrhein.de}
}

\maketitle
\conf{\textit{  Proc. of International Conference on Artificial Intelligence, Computer, Data Sciences and Applications (ACDSA 2024) \\ 
1-2 February 2024, Victoria-Seychelles}}
\begin{abstract}
 Probabilistic world models increase data efficiency of \ac{MBRL} by guiding the policy with their epistemic uncertainty to improve exploration and acquire new samples. Moreover, the uncertainty-aware learning procedures in probabilistic approaches lead to robust policies that are less sensitive to noisy observations compared to uncertainty-unaware solutions. We propose to combine trajectory sampling and \ac{DGCN} for a data-efficient solution to \ac{MBRL} problems in an optimal control setting. We compare trajectory sampling with density-based approximation for uncertainty propagation using three different probabilistic world models; \acp{GP}, \acp{BNN}, and \acp{DGCN}. We provide empirical evidence using four different well-known test environments, that our method improves the sample-efficiency over other combinations of uncertainty propagation methods and probabilistic models. During our tests, we place particular emphasis on the robustness of the learned policies with respect to noisy initial states.
\end{abstract}



\def\thefootnote{*}\footnotetext{Equal contribution, alphabetical order}
\def\thefootnote{$\dagger$}\footnotetext{Work done while at Niederrhein University }
\def\thefootnote{\arabic{footnote}}
\section{Introduction and Related Work}\label{sec1}

\Ac{MBRL} is a popular approach, especially for solving problems with high sample costs. For example, the use of a learned world model is very successful in image-based environments \cite{HS18}. Such a world model enables training an agent using a model-free RL algorithm such as \ac{PPO} \cite{SWDRK17} and to evaluate the agent in the real world afterwards \cite{Atari19}. In many cases, the dynamics model is incorporated directly into an objective function to maximize the expected cumulative reward for the long-term planning of a task. Probabilistic models are the foundation of most data-efficient methods in \ac{MBRL}, since these models are able to incorporate aleatoric (due to randomness/noisy observations) and epistemic (due to lack of knowledge) uncertainty into long-term planning. They allow to choose actions that maximize the expected reward and avoid high risk or low performance behaviour. Moreover, it is possible to enforce exploration by using the epistemic uncertainty of the model. The main idea of probabilistic approaches such as \ac{PILCO} \cite{Deis10, DR11, DFR13}, Deep \ac{PILCO} \cite{GAR16}, DGP-MPC \cite{GHLK20} and \ac{PETS} \cite{CCAL18} is to define states and possibly actions with a probability distribution instead of a deterministic value and to propagate the uncertainty iteratively through the model. Therefore, the expected cumulative reward is often maximized using fewer trials compared to model-free strategies. Common models are \ac{GP} \cite{RW06} and \ac{BNN} \cite{Neal1995}. The probabilistic \ac{MBRL} approaches have been successfully applied to tasks with limited complexity and in particular, to applications in robotics. A detailed overview is given by \cite{DNP13,MBJJ20}.

Generally, a distinction between policy-based and policy-free methods can be made. For example, the original \ac{PILCO} algorithm as well as Deep \ac{PILCO} use a parametrized policy, but \ac{PILCO} has been modified to a policy-free method with \ac{MPC} \cite{KD18}. \Ac{PETS} and DGP-MPC also follow the \ac{MPC} approach. On one hand, a learnt policy represented as a function is often more compute-efficient after training, but it is also limited in its capabilities, since it was trained to solve a specific task in a fixed setting. On the other hand, \ac{MPC} is more flexible and promising in environments with changing requirements, i.e. where the target state is not clearly defined, but it involves an open loop optimization problem during application. \change{In this paper, we focus on applying the trajectory sampling as proposed by \ac{PETS} to policy-based applications}. Furthermore, this approach also increases the comparability of all methods. 

\ac{PILCO} is based on a \ac{GP} with squared exponential kernel, which can be seen as a limited dynamics model as it is found to be unrealistically smooth for real world applications \cite{RW06,Stein1999}. However, Deep \ac{PILCO} and \ac{PETS} are based on \acp{BNN}. The authors argue that the uncertainty quantification capabilities as well as the scalability of their models are of particular importance. We use \ac{DGCN} \cite{Crem21} as a recently published alternative, which overcomes the shortcomings of a simple \ac{GP} model. In particular, we introduce \ac{DGCNTS}. Our goal is to investigate the most sample-efficient combination of uncertainty propagation method and probabilistic model for policy search. 

\section{Probabilistic Models}\label{sec2}

\subsection{Gaussian Process Regression}\label{secGP}

Formally, the \ac{GP} posterior is given as \cite{RW06}
\change{\begin{equation}
	\hspace*{-0.05cm}
	\begin{aligned}
		&\!\detv{f}^* \sim \mathcal{N}\left(\mu(\detm{X}^*), \sigma^2(\detm{X}^*)\right) \\
		&\!\mu(\detm{X}^*)\!=\!K(\detm{X}^*, \detm{X}) K(\detm{X}, \detm{X})^{-1} \detv{f} \\
		&\!\sigma^2(\detm{X}^*)\!=\!K(\detm{X}^*, \detm{X}^*)\!-\!K(\detm{X}^*, \detm{X}) K(\detm{X}, \detm{X})^{-1}\!K(\detm{X}, \detm{X}^*)
	\end{aligned}
\end{equation}
where $\detm{X}^*$} denotes the input coordinates of the test data, $\detv{f}^*$ the corresponding function values, $\detm{X}$, $\detv{f}$ denote the training data and $K$ the covariance function. Specifically, the element $K(\detm{X}^*, \detm{X})_{i,j} = k(\detv{x}^*_{i}, \detv{x}_j)$ is computed using the kernel function $k$ and the corresponding entries $\detv{x}^*_{i}, \detv{x}_j$ of the input points. Some of the most popular \ac{GP} kernels are distance-based and stationary such as the squared exponential kernel
\begin{equation}
	k(\bar{r}) = \exp\left(-\frac{\bar{r}^2}{2}\right)
\end{equation}
where $\bar{r}$ is the distance between two points $\detv{x}_i, \detv{x}_j$, scaled by some positive kernel parameter $\theta_l$, often called the \textit{length-scale}, \ie
\begin{equation}
	\bar{r} = \Big\|\frac{\detv{x}_i - \detv{x}_j}{\theta_l}\Big\|
\end{equation}
whereas for anisotropic \ac{GP} it is used a length scale $\theta_{l,1}, \dots \theta_{l, d}$ for each of the $d$ dimensions.

\subsection{Deep Gaussian Covariance Network}\label{secDGCN}

\begin{figure*}
	\begin{center}
	\includegraphics[width=.6\textwidth]{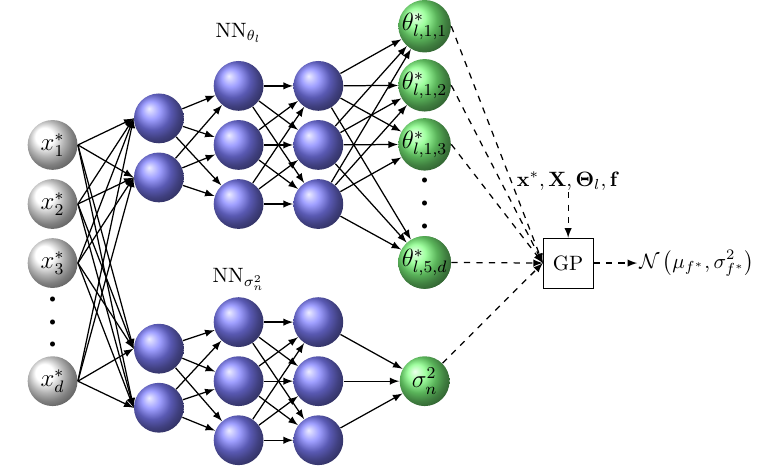}
	\end{center}
	\caption{Schematic overview of \ac{DGCN}. $\boldsymbol{\Theta}_l$ represents the matrix of length scales for all training points $\detm{X}$, as output by the \ac{NN}.}\label{fig:DGCN}
	\vspace*{-.3cm}
\end{figure*}

A \ac{DGCN} \cite{Crem21} is an instationary \ac{GP} model, using an \ac{NN} 
to estimate the kernel parameters depending on the prediction point. Similarly to a \ac{GP} model, \Ac{DGCN} also deploys distance-based kernels, specifically the squared exponential kernel, three Matérn kernels with the parameters $\nu \in {0.5, 1.5, 2.5}$ as well as the rational quadratic kernel (see \cite{RW06}). However, the globally constant $\theta_l$ is replaced for each kernel by a local prediction $\theta_l(\cdot)$, which is the output of an \ac{NN}. Thus, the scaled distance is computed as
which can be extended trivially to the anisotropic case by increasing the number of \ac{NN} outputs, where each dimension is assigned a separate length scale, \ie an \ac{NN} output (see Figure \ref{fig:DGCN}). Similar to obtaining $\theta_l$ for \ac{GP}, the \ac{NN} is optimized by maximizing the Type-II likelihood as given in \cite{RW06}. 

The non-stationary definition of the distance based kernels allows an automated kernel selection even for heteroscedastic data, for which the optimal choice of kernel may change in different subregions. Moreover, an additional network output parametrizes the assumed noise for $K(\detv{x}, \detv{x})$ ($\sigma_n$ in \cite{RW06}), which allows handling heteroscedastic uncertainty, i.e. the noise level depends on $\detv{x}$.

Finally, batch training can be used to obtain the network weights, which accelerates and stabilizes learning from larger data sets compared to vanilla \ac{GP}. See chapter 2.10 in \cite{Crem21} for further details.

\subsection{Probabilistic Neural Networks}\label{secPNN}
\Acp{PNN} aim to find a predictive distribution parametrized by the network outputs similar to the \acp{GP}. Thus, they can be described as \acp{BNN} \cite{Neal1995}. Nevertheless, they are not full Bayesian models, since the weights have fixed values after the training. In this work, the definition in \cite{CCAL18} is used, which assumes a Gaussian posterior. As such, the predictive loss function is proportional to the Gaussian log-likelihood
\begin{equation}
	\mathcal{L} = \frac{\left(\mu_f - f\right)^2}{\sigma_f} + \log\left(\sigma_f\right)
\end{equation}
for each observed function value $f$, where $\mu_f,\sigma_f$ are the outputs of an \ac{NN}, parametrizing the posterior. Moreover, weight decay is used since the deployed networks are often overparametrized. 

Although \acp{PNN} deliver uncertainty estimates, authors in \cite{CCAL18} argue that this is not sufficient and propose to use ensembles for quantifying the epistemic model uncertainty, similar to \cite{Lakshminarayanan2017}. This formulation brings \acp{PNN} closer to a full Bayesian model, as multiple sets of weights are used for prediction. As such, both $\mu_f$ and $\sigma_f$ are computed from the arithmetic mean of individual networks in the ensemble \change{(E-\ac{PNN})
\begin{equation}
		\mu_f = \frac{1}{m}\sum_i^m  \mu_{f,i},  \quad \sigma_f = \frac{1}{m}\sum_i^m  \sigma_{f,i}, \label{eq:epnn}
\end{equation}
where $\mu_{f,i},\sigma_{f,i}$ denote the outputs for the $i$-th set of \ac{NN} weights, $m$ denotes the total number of weight sets, or equivalently the number of \acp{NN} in the ensemble.}

The importance of the provided heteroscedastic uncertainty estimates by the E-\ac{PNN} model is emphasized in \cite{CCAL18}; aleatoric as well as epistemic. As explained in Section \ref{secDGCN}, \ac{DGCN} exhibits similar properties, which makes the comparison of these two models particularly interesting.

\section{Propagation of Uncertainty}\label{sec3}

In the following, we denote the state as $s_t \in S$ and the action as $a_t \in A$, where $S$ denotes the state space, $A$ the action space and $t$ the time index. A trial yields state-action inputs $(s_t, a_t)$ and the corresponding output pairs $(s_{t+1}, r_t)$ for $t=0, \dots, T-1$, where $T$ is the time horizon and $r_t$ is the immediate reward for transitioning from $s_t$ to $s_{t+1}$ by taking the action $a_t$. 

In the case of continuous state spaces, it is common to use a regression model to predict the subsequent state $s_{t+1}$ from the current state-action pair $(s_t, a_t)$. More precisely, the collected data is used to model the underlying functional relation
\begin{align} s_{t+1} = s_t + f(s_t, a_t) \label{dynamics} 
\end{align}
of the dynamics, where $f$ possibly contains a zero-mean noise term, representing the aleatoric uncertainty.

If $f$ is learned by a probabilistic model, $f(s_t, a_t)$ contains aleatoric as well as epistemic uncertainty. For the models discussed in Section \ref{sec2}, $f(s_t, a_t)$ follows Gaussian distribution. Propagation of uncertainty refers to the situation, where the input itself is a probability distribution. Hence, in the present setting, the state-action pair $(s_t, a_t)$ as well as the transition function $f$ entail some randomness. Therefore, the question arises how to determine the uncertainty of \change{the transition value $\Delta_t$} and how to incorporate it into policy search. \change{We reformulate equation (\ref{dynamics}) in the form
\begin{align} \Delta_t : = s_{t+1} - s_t = f(s_t, a_t) = f(s_t, \pi_{\theta}(s_t)),
\end{align}
where $\pi_{\theta}$ denotes the policy with parameters $\theta$.}

\subsection{Density-based uncertainty propagation} \label{density}

The distribution of $\Delta_t$ is given by
\begin{align} p(\Delta_t)= \int p(f(s_t, \pi_{\theta}(s_t))~\mid~s_t)~p(s_t) ds_t,  \label{pDelta}
\end{align}

where $p(s_t)$ denotes the state distribution at time $t$. Unfortunately, it is not possible to compute this integral in a closed form. Usually, it is assumed that $s_t$ admits a Gaussian distribution and $p(\Delta_t)$ is approximated by another Gaussian. If the mean and variance of the distribution of $\Delta_t$ as well as the input-output covariance is known, it is possible to compute a Gaussian distribution for $s_{t+1}$.

The \ac{PILCO} algorithm assumes that $f$ is given by a \ac{GP} with squared exponential kernel. In this case, the mean and variance can be computed explicitly using closed-form equations. This method is referred to as moment matching. Although mathematically pleasant, it is quite restricting regarding the model, the policy as well as the reward function family. Alternatively, methods such as \ac{EKF}, \ac{UKF} and particle methods can be used for approximation \cite{KF09}, which also allow a general class of probabilistic models, policies and rewards. In particular, Deep \ac{PILCO} uses a \ac{BNN} instead of a \ac{GP} in combination with a particle method to calculate the parameters of a Gaussian distribution. More precisely, sample statistics of the predictions are used for the approximation. However, these alternative methods perform possibly worse compared to exact moment matching \cite{DFR13}. 

Since we are interested in a comparison with Deep \ac{PILCO} as well, we use a suitable particle method for the models presented in Section \ref{sec2}. We calculate the mean of output distribution by
\begin{align}
	\mu(\Delta_t) &= \frac{1}{n_p} \sum_{q=1}^{n_p} \mu(s_t^q) \label{PFmean}
\end{align}
and the covariance \change{by
\begin{equation}
	\begin{aligned}
		\Sigma(\Delta_t) &= \frac{1}{n_p} \sum_{q=1}^{n_p} \Sigma(s_t^q) + \Delta_\mu\Delta_\mu^T \\ \label{PFvar}
		\Delta_\mu &= \mu(s_t^q) - \mu(\Delta_t),
	\end{aligned}
\end{equation}
where} $n_p$ denotes the number of particles and $\mu(s_t^q)$ and $\Sigma(s_t^q)$ are mean prediction and covariance of the $q$-th particle, respectively. In particular, the particles $s_t^q$, $q=1,\dots,n_p$ are sampled from the Gaussian distribution of $s_t$. Note that eqs. (\ref{PFmean}) and (\ref{PFvar}) are straightforward Monte-Carlo approximations of the mean and variance of the distribution given in eq. (\ref{pDelta}). Similarly, the input-output covariance can be estimated from the sample statistics. Density-based uncertainty propagation with a pre-trained policy and \ac{PILCO} is illustrated in Figure \ref{fig4}. Please refer to \cite{Deis10} for further details on density-based uncertainty propagation.

\subsection{Trajectory sampling}

Please note that the actual distribution in eq. (\ref{pDelta}) could indeed be a multimodal distribution and a Gaussian distribution is possibly a bad approximation. For this reason, the \ac{PETS} algorithm proposes a different strategy. A fixed number of particles is sampled from the initial distribution of $s_0$ and the probabilistic model predicts each particle separately, i.e., a particle $s_t^q$ is propagated by predicting $f(s_t^q, a_t)$ and sampling $s_{t+1}^q$ from the output distribution, where $q$ denotes the number of the particle. In this way, each particle yields a trajectory. By averaging the cumulative reward over all trajectories, the expected cumulative reward can be estimated. This method is highly flexible and avoids any issues due to the unimodal approximation. For further details please refer to \cite{CCAL18}. The results of trajectory sampling with a pre-trained policy as well as \ac{DGCN} and an ensemble of \ac{PNN}s are illustrated in Figures \ref{fig5} and \ref{fig6}, respectively.

\section{Trajectory Sampling with Deep Gaussian Covariance Networks}\label{sec4}

Algorithm \ref{algo1} describes the proposed \ac{DGCNTS} method, which is a \ac{DGCN}-based \ac{MBRL} framework using trajectory sampling. Note that $\pi_{\theta}$ and the reward could also be time-dependent (non-stationary) and the involved distributions do not necessarily need to be Gaussian. 

\begin{algorithm}[H]
	\caption{DGCNTS}\label{algo1}
	\begin{algorithmic}[1]
		\State \change{Collect initial data $\mathcal{D}$ with a random policy}
		\State Define policy $\pi_{\theta}: S \rightarrow A$ with parameters $\theta$
		\State Initialize $\theta$ randomly
		\For{trial $i = 1, \dots, n_r$}
		\State Fit \ac{DGCN} dynamics model $f$ given $\mathcal{D}$
		\Repeat
		\State Sample initial states $s^q_0$, $q=1,\dots, n_p$, from $p(s_0)$
		\For{time $t = 0, \dots, T-1$}
		\For{$q=1,\dots,n_p$}
		\State Compute action $a^q_t = \pi_{\theta}(s^q_t)$
		\State \change{Predict $f(s^q_t, a^q_t) \sim \mathcal{N}(\mu(s^q_t), \Sigma(s^q_t))$}
		\State Sample $\Delta^q_t$ from  $\mathcal{N}(\mu(s^q_t), \Sigma(s^q_t))$
		\State Set $s^q_{t+1} = s^q_t + \Delta^q_t$
		\State Compute reward $r_t^q$
		\EndFor
		\State Compute average reward $r_t = \frac{1}{n_p} \sum_{q=1}^{n_p} r_t^q$
		\EndFor
		\State Calculate expected cumulative cost as $-\sum_{t=0}^{T-1} r_t$
		\State Apply optimizer and update $\theta$
		\Until{Optimizer converges}
		\State Perform trial (rollout) and collect data $\mathcal{D}_i$
		\State Add $\mathcal{D}_i$ to $\mathcal{D}$
		\EndFor
	\end{algorithmic}
\end{algorithm}

There are two important differences between \ac{DGCNTS} and \ac{PETS}, besides the model choice. Firstly, \ac{DGCNTS} does not deploy an ensemble, as \ac{DGCN} as a \ac{GP} models the epistemic uncertainty naturally. Secondly, \ac{DGCNTS} seeks to find a policy during training instead of using an \ac{MPC} strategy as described before. Nonetheless, the second point is purely a design choice, which is why we test the model in \ac{PETS} within the same framework. The application of \ac{MPC} may be part of future work.

\section{Experimental results}\label{sec5}

\begin{figure*}[!htb]%
	\centering
	\begin{subfigure}{.48\linewidth}
		\centering
		\includegraphics[width=\txtw\textwidth]{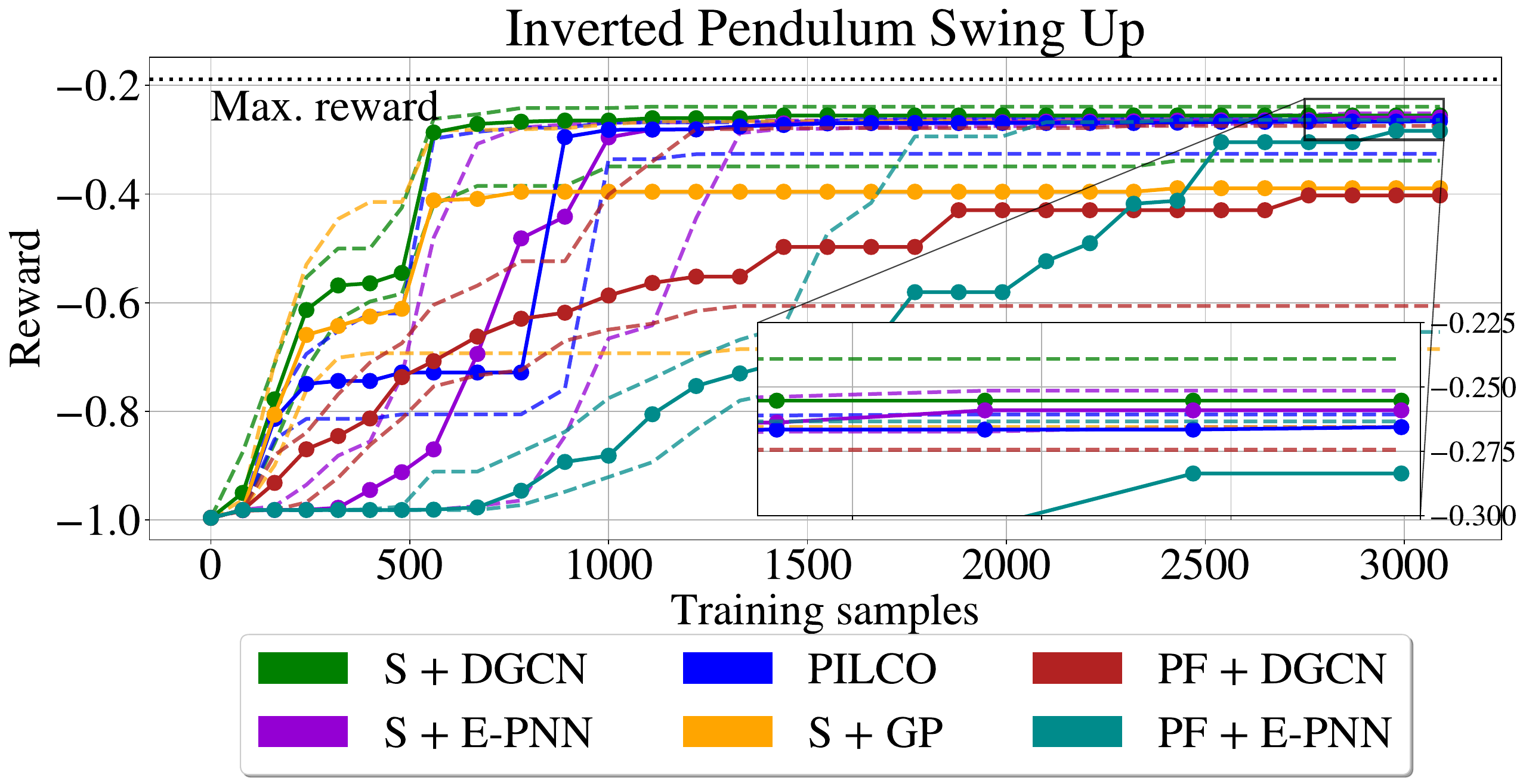}
		\caption{Results for \ac{IPSU} task}\label{fig2}
	\end{subfigure}
	\begin{subfigure}{.48\linewidth}
		\centering
		\includegraphics[width=\txtw\textwidth]{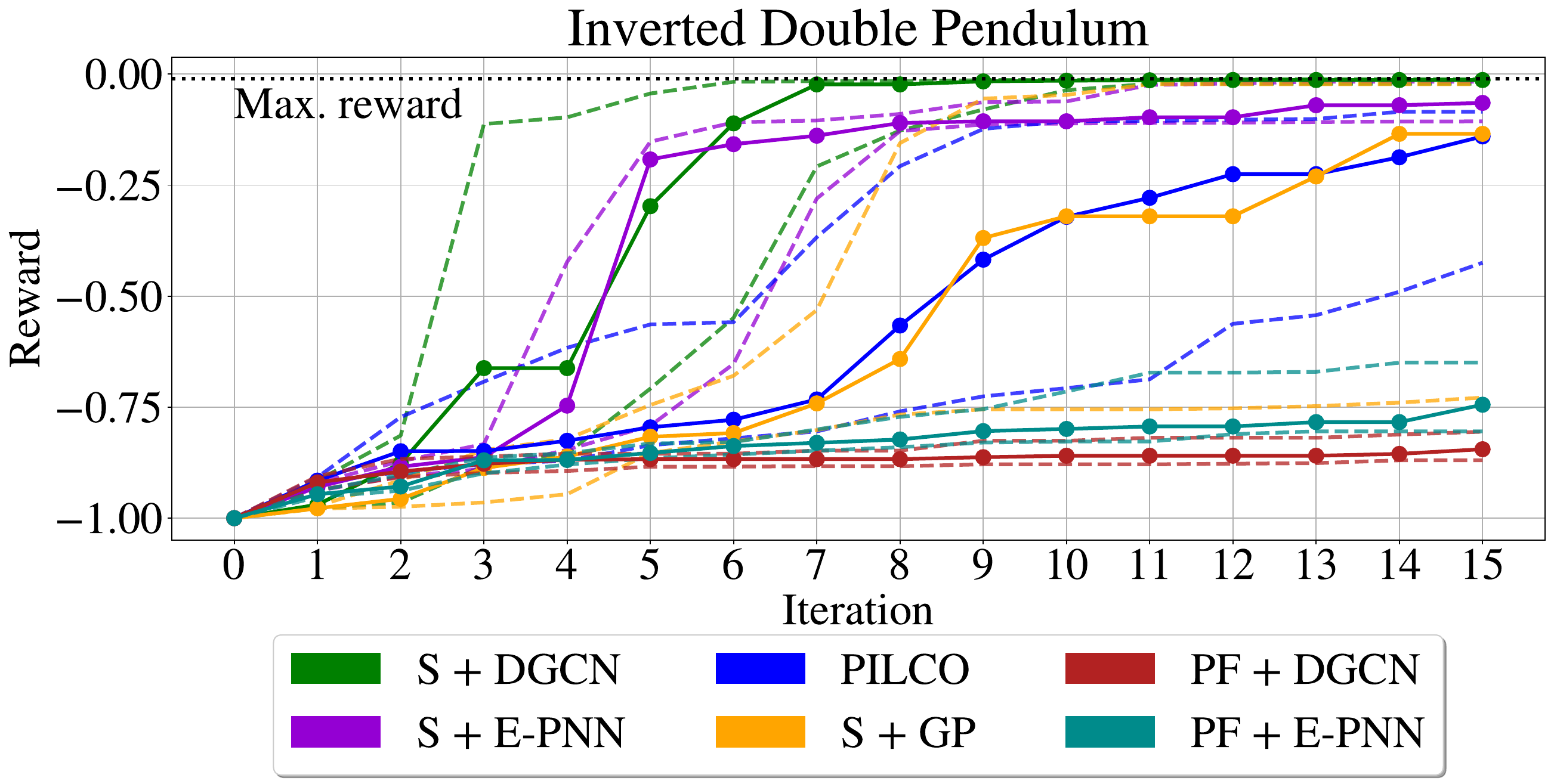}
		\caption{Results for \ac{IDP} task}\label{fig3}
	\end{subfigure} \\ \vspace{.15cm}
	\begin{subfigure}{.48\linewidth}
		\centering
		\includegraphics[width=\txtw\textwidth]{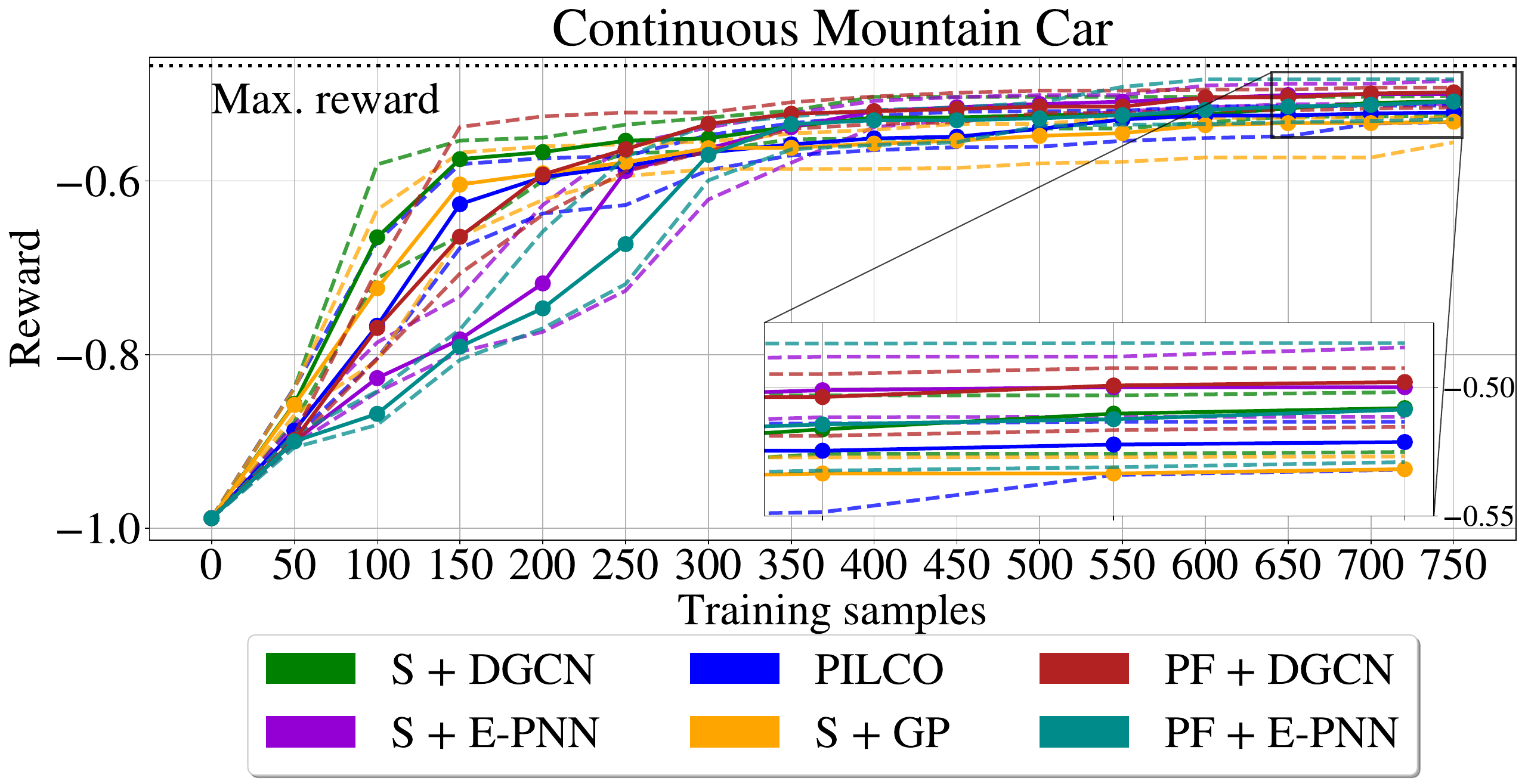}
		\caption{Results for \ac{CMC} task}\label{fig8}
	\end{subfigure}
	\begin{subfigure}{.48\linewidth}
		\centering
		\includegraphics[width=\txtw\textwidth]{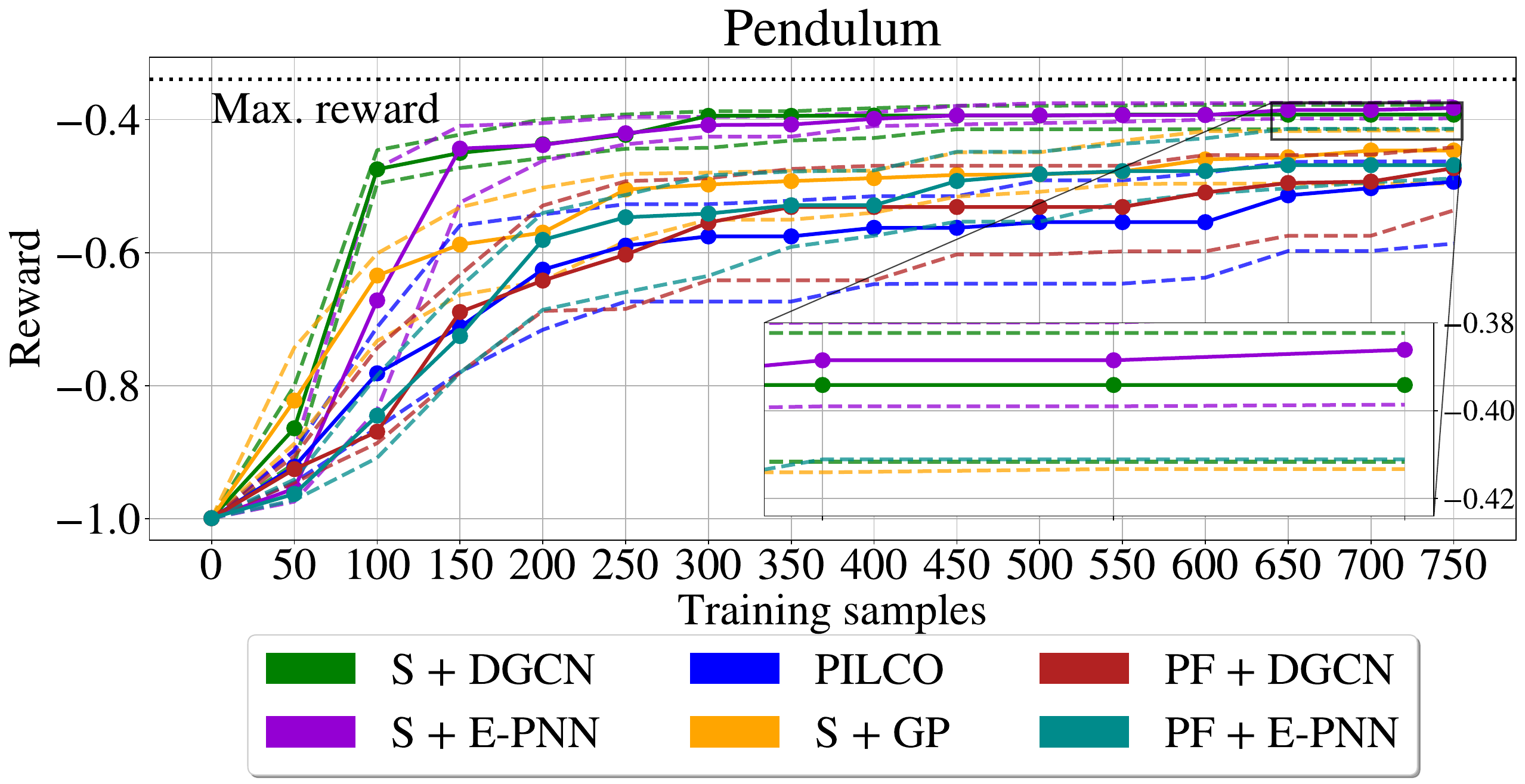}
		\caption{Results for \ac{P} task}\label{fig9}
	\end{subfigure}
	\caption{Results of benchmarked tasks}
\vspace*{-.3cm}
\end{figure*}

The following noise-free tasks are investigated using the corresponding gym environments \cite{gym16}: the \ac{IPSU} task, the \ac{CMC} problem, the \ac{IDP} and the \ac{P} environments. \Ac{IPSU} and \ac{IDP} are simulated using MuJoCo \cite{mujoco12} and PyBullet \cite{pybullet, benelot} physics engines, respectively. A detailed discussion of the experiments with these environments is given in the following\footnote{The code is available at \url{\github}.}. 

Each task requires to reach and maintain a specific target state; for example, reaching and holding an upright position of the pendulum. All experiments are performed with an \acs{RBF} policy as described in \cite{DFR13} and a task dependent number of basis functions. The actions are bounded by the trapezoidal squashing function presented in \cite{DFR13}. The reward $r_t$ is computed as
\begin{align} r_t = \exp\big(- (s_{t+1} - s_{\text{tar}})^T W (s_{t+1} - s_{\text{tar}})\big), \label{expreward}
\end{align}
where $s_{\text{tar}}$ denotes the target state and $W$ is a diagonal matrix which attaches a weight to each component of the target. 

The reward function in eq. (\ref{expreward}) is also called exponential reward and takes values in $(-1, 0]$. Sampling based methods are able to approximate the expectation of arbitrary reward functions, but \ac{PILCO} is limited to functions, which admit an analytic formula for its expectation with respect to a Gaussian distributions. Hence, we restrict ourselves to rewards of the form given in eq. (\ref{expreward}) for comparability. The \ac{IDP} task differs from others in that a trial terminates if the angle of the pendulum to the vertical axis exceed a given threshold. In this case, all rewards related to remaining time steps is set to $-1$ in order to reflect the unsuccessful termination. Optimization of the policies was performed using the Adam optimizer \cite{KB14} with multiple restarts and early stopping. Further details on the used hardware is given in Section \ref{setup} and the detailed experimental setup can be found in Table \ref{table1}. Moreover, the computation times are illustrated in Figure \ref{fig7}.

Since the approaches at hand take the probability distributions into account explicitly and in particular, uncertainties with respect to the initial state $s_0$, we expect that the learned policies possess robustness regarding perturbations in $s_0$. For this reason, each policy is tested on 20 random initial states and its performance is measured by the average reward. The initial states are sampled once and used for each policy independent of the search strategy. 

Furthermore, each experiment was repeated $20$ times with random initializations of the policy, environments and models. The resulting figures show the median as well as the $25\%$ and $75\%$ quantiles of the outcomes after each iteration. Note that the dotted horizontal line in the figures below is the result of numerical optimization of the objective function. Therefore, this information is only a reference point for the true maximal reward and required much more experiments than the investigated algorithms. 

In the following, we use the abbreviations S for trajectory sampling and PF for the particle method explained in Section \ref{density}. Specifically, S + \ac{DGCN} equals the \ac{DGCNTS} algorithm stated in Section \ref{sec4}. Moreover, E-\ac{PNN} is an ensemble of $m=5$ (see Eq. \ref{eq:epnn}) \ac{PNN}s and the computations are based on the implementation available from \ac{PETS} \cite{CCAL18}. For \ac{GP}, only the squared exponential kernel is used for the applicability of the moment matching method, as proposed in \ac{PILCO}. Here, we used a GPflow implementation \cite{gpflow16}.

The figures for \ac{IPSU}, \ac{CMC} and \ac{P} use the used number of training samples for the dynamics model on the horizontal axis in order to emphasize how much data was needed to achieve the corresponding reward. For these examples, the number of samples can also be directly converted into the number of performed trials (see information in Table \ref{table1}). However, we solely use the trials for \ac{IDP} due to the environment termination condition discussed earlier. The same amount of samples could be obtained from much more experiments, but we like to measure data-efficiency mainly by the required number of trials.

Figure \ref{fig2} shows that S + \ac{DGCN}, \ac{PILCO} and S + E-\ac{PNN} yield the best results, but S + \ac{DGCN} is clearly the most data-efficient method. In the end, PF + E-\ac{PNN} is able to provide almost equivalent results, but the policy improves very slowly. The \ac{IDP} example illustrated in Figure \ref{fig3} requires fast learning to collect a decent amount of data, since bad performance of the policy is additionally penalized by termination of the environment. Both models using the particle method are not able to solve the task and both approaches using the vanilla \ac{GP} show moderate results. However, \ac{DGCN} and E-\ac{PNN} are able to find a good policy in a data-efficient way. The highest reward was achieved by \ac{DGCN} in combination with trajectory sampling (\ac{DGCNTS}).

Almost all approaches perform equally good in the \ac{CMC} task presented in Figure \ref{fig8}. However, the two models using the density-based particle approach (PF + \ac{DGCN} and PF + E-\ac{PNN}) trail behind. Since both models perform very well in combination with trajectory sampling for all examples, we reason that this observation is rather caused by the uncertainty propagation method.

The illustration of the results for the \ac{P} task in Figure \ref{fig9} indicates once again that trajectory sampling yields the best results. S + \ac{DGCN} as well as S + E-\ac{PNN} are very comparable, whereas the four remaining strategies perform worse.

\section{Conclusion and Outlook}\label{sec6}

The experimental results clearly show that trajectory sampling in combination with \ac{DGCN} or E-\ac{PNN} performed better than the density-based uncertainty propagation methods for the investigated tasks and the difference to other strategies was especially large in \ac{IDP} and \ac{P} environments. A probable explanation is that the propagated distributions are poorly approximated using a Gaussian. We investigated this problem in Appendix \ref{uncertainty}. For this reason, we agree with the arguments given in \cite{CCAL18}; density-based uncertainty propagation as used in \ac{PILCO} may be inappropriate in case of non-linear dynamics, whereas trajectory sampling is still applicable. 

Additionally, the proposed method is highly flexible and a trade-off between the accuracy of uncertainty propagation method and the computation time is easily possible by adjusting of the number of particles. Moreover, all examples indicate that our \ac{DGCNTS} algorithm achieved good results with few samples. In this regard, \ac{DGCNTS} even outperforms the combination of trajectory sampling and E-\ac{PNN}, a result contradicting the general conjectures about \ac{GP} in the original \ac{PETS} paper. Although not a vanilla one, \ac{DGCN} is also a \ac{GP} and it clearly achieved comparable or better results compared to E-\ac{PNN} in comparable time.

Currently, our results are limited to the policy-based setting and to tasks with limited complexity. In future work, the proposed framework should be extended to more complex tasks involving \ac{MPC} as well as higher dimensional state and (possibly discrete) action spaces. In particular, the scalability of \ac{DGCN} regarding the number of samples is of special interest. Furthermore, we did not focus on exploration is the early stage of learning which could yield an additional benefit in terms of data-efficiency and performance.




%

\printbibliography

\appendix

\section{Appendix}\label{appendix}

\subsection{Uncertainty propagation} \label{uncertainty}

\begin{figure*}[!htb]%
	\begin{subfigure}[T]{0.33\linewidth}
		\centering
		\includegraphics[width=\txtw\textwidth]{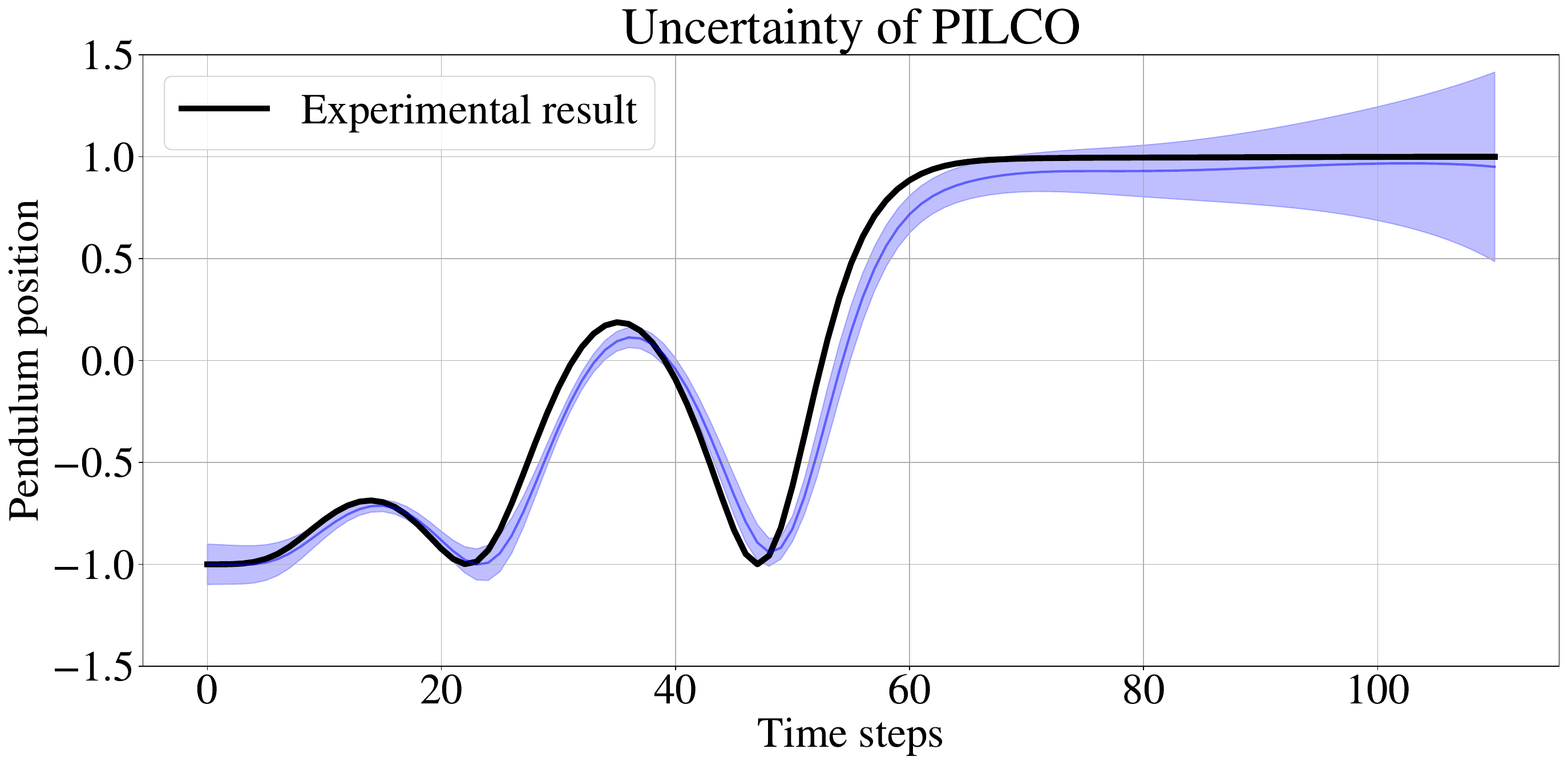}
		\caption{Uncertainty of \ac{PILCO} after 12 iterations}\label{fig4}
	\end{subfigure}
	\begin{subfigure}[T]{0.33\linewidth}
		\centering
		\includegraphics[width=\txtw\textwidth]{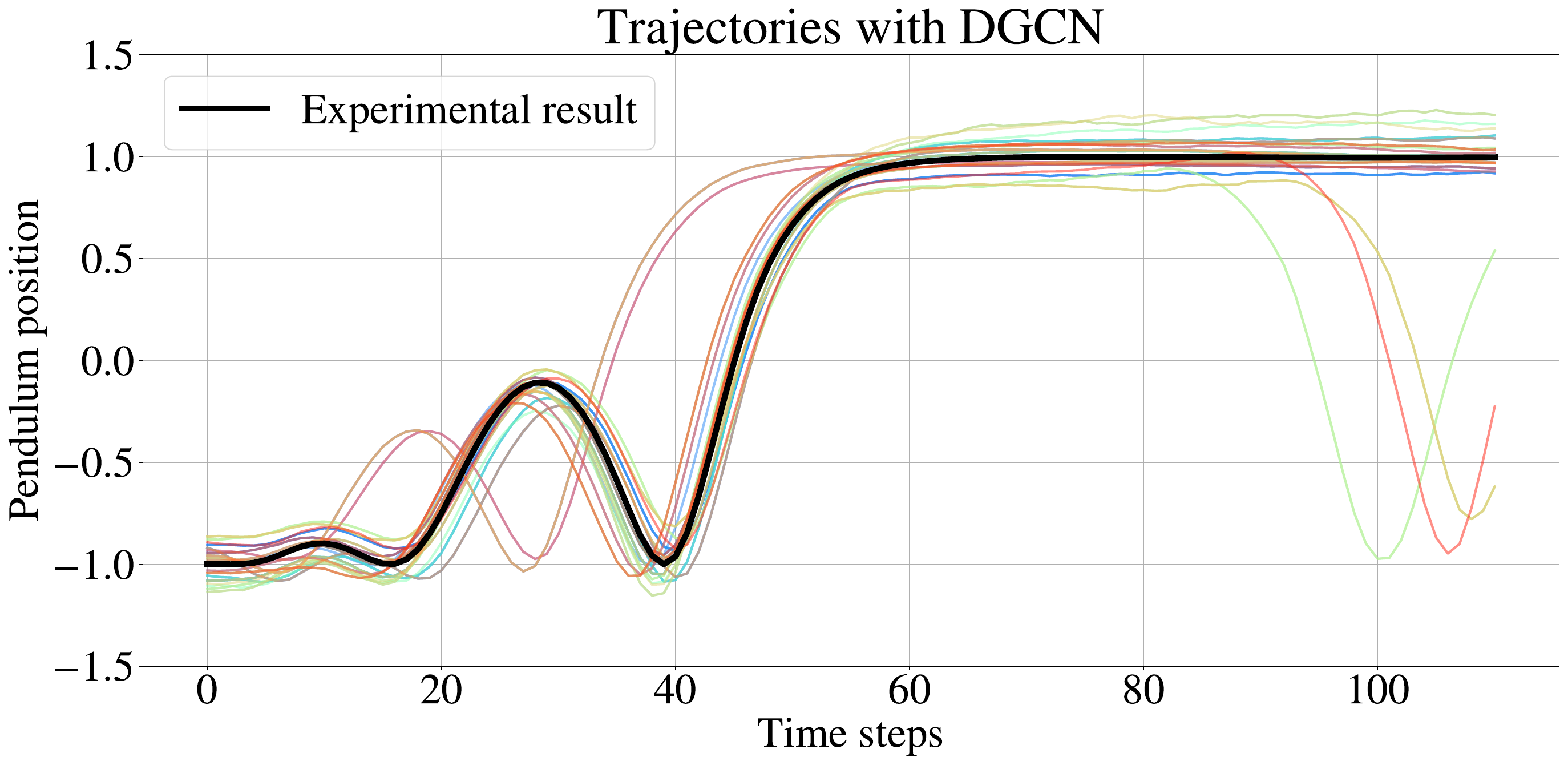}
		\caption{Sampled trajectories with \ac{DGCN} after 12 iterations}\label{fig5}
	\end{subfigure}
	\begin{subfigure}[T]{0.33\linewidth}
		\centering
		\includegraphics[width=\txtw\textwidth]{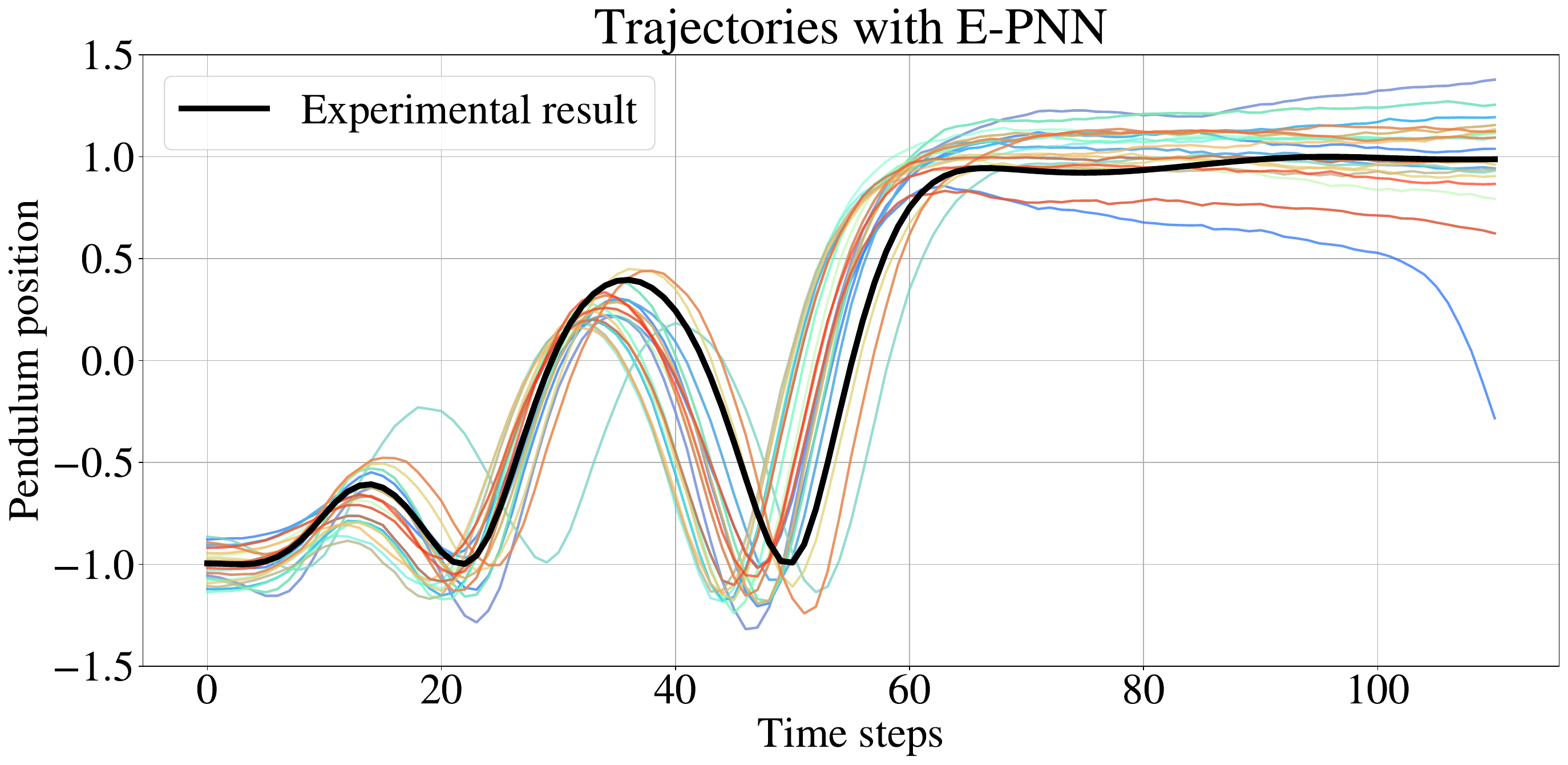}
		\caption{Sampled trajectories with \ac{PNN} after 12 iterations}\label{fig6}
	\end{subfigure}
\caption{Epistemic uncertainty of the tested models at \ac{IPSU} task after 12 iterations}
\vspace*{-.3cm}
\end{figure*}

Figure \ref{fig4} shows the model prediction of \ac{PILCO} in combination with a pre-trained policy. The pendulum position refers to the vertical position of the pendulum in the \ac{IPSU} task. Thus, the actual position can take values between $-1$ and $1$. The blue line illustrates the mean prediction of the algorithm, whereas the surrounding light blue area depicts the 90$\%$ credible interval. Figure \ref{fig5} is similar to Figure \ref{fig4}, but uses \ac{DGCN} with trajectory sampling instead of \ac{PILCO}. Each colored line refers to one possible trajectory of the dynamics. Finally, Figure \ref{fig6} illustrates trajectory sampling with an ensemble of \acs{PNN}.

Please note that trajectories which have large distance to the experimental result do not necessarily indicate a bad model, since they possibly arise from different initial states $s_0$ and therefore, the dynamical system shows a different behaviour. However, in some cases the model proposes values outside the physical reasonable range between $-1$ and $1$. Moreover, Figure \ref{fig5} and Figure \ref{fig6} already indicate that the Gaussian distribution yields in some cases a poor approximation of the distribution of states at a fixed time step. This assumption is encouraged by Figure \ref{fig65}, which shows the actual distribution of the pendulum position after one time step using 200.0000 samples:

\begin{figure}
	\centering
	\begin{minipage}{0.99\linewidth}
		\centering
		\includegraphics[width=\txtw\textwidth]{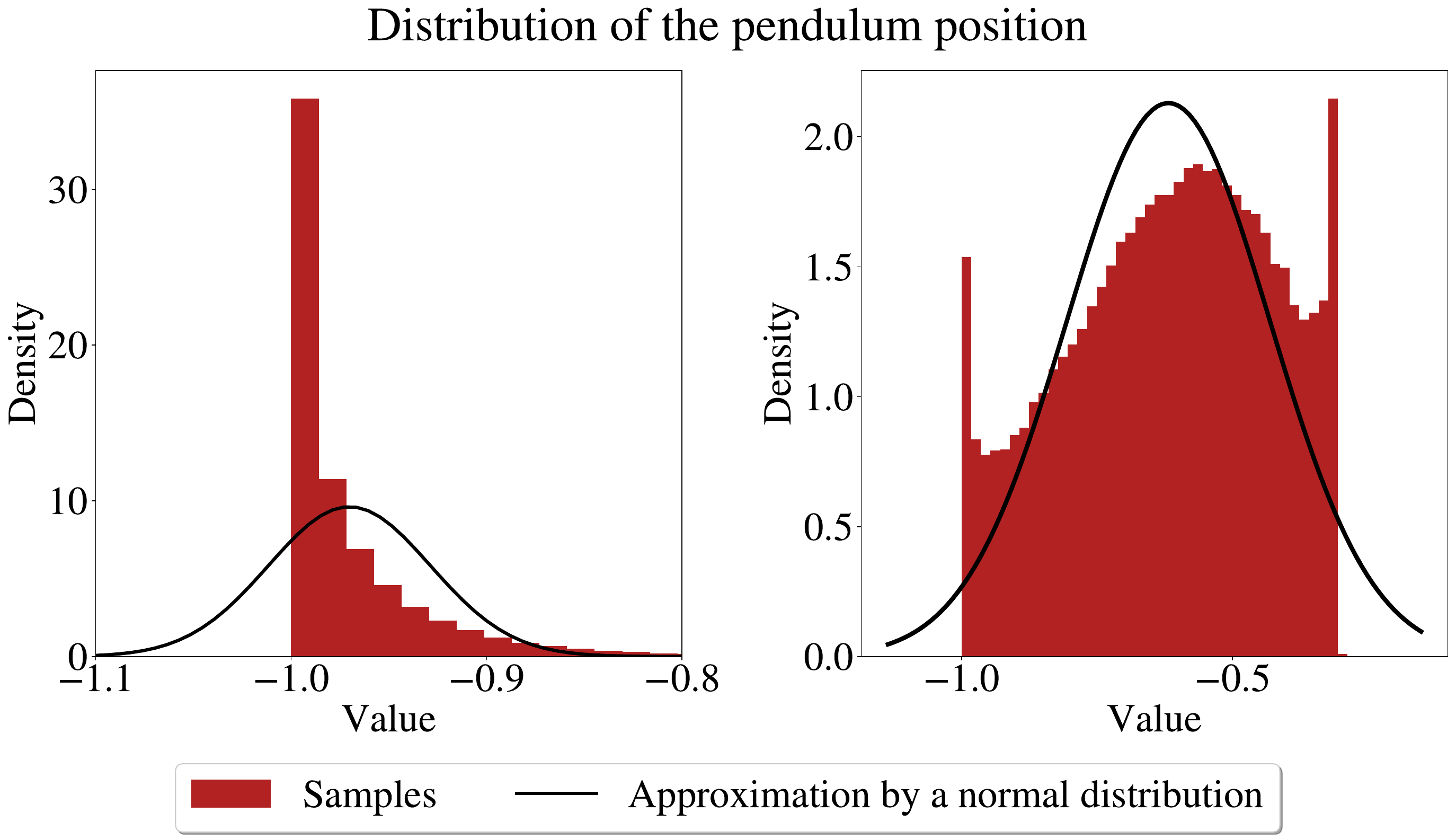}
		\captionof{figure}{Asymmetric unimodal and multimodal distributions observed during \ac{IPSU} task}\label{fig65}
	\end{minipage}
\end{figure}

\subsection{Experimental setup} \label{setup}

The experiments were performed on an internal GPU cluster equipped with four NVIDIA A100 GPUs, two AMD EPYC 7662 CPUs and 1024 GB memory per node. 16 experiments were performed in parallel for the tasks \ac{CMC}, \ac{P} and \ac{IDP}. Only 4 experiments in parallel were executed for the task \ac{IPSU}, since the memory demand is higher due to a longer planning horizon, more samples as well as a higher number of iterations. Moreover, due to the computational complexity of the \ac{PILCO} algorithm and the memory limits, it was necessary to limit the model training data to 800 samples. For this purpose, the latest 800 samples collected from previous trials are used for model training. This approach only affects the \ac{IPSU} task, since the limit of 800 samples is reached after 9 iterations. The remaining examples do not exceed this limit before the last trial. Alternatively, the sample size could for example be reduced by using a sparse GP \cite{Titsias09} as proposed in \cite{Deis10}, but the modification of the model also affect (the equations of) the \ac{PILCO} algorithm itself.

\begin{table}
	\centering
	\setlength\tabcolsep{3pt}
	\captionsetup{justification=centering}
	\caption{Experimental setup}
	\begin{tabular}{lcccc}\toprule
		& CMC & P & IDP & IPSU \\\midrule
		State dimensions & 2 & 3 & 6 & 5 \\
		Initial samples & 50 & 50 & 54 & 80 \\
		Learning horizon$^*$ & 50 & 50 & \hspace{0.15cm} 50$^{**}$ & 80/110 \\
		Evaluation horizon$^{***}$  & 50 & 50 & 200 & 200 \\
		Basis functions &  35 & 35 & 40 & 100 \\
		Trainable parameters & 107 & 143 & 286 & 605 \\
		Trajectories & 100 & 100 & 100 & 50 \\
		Iterations   & 15 & 15 & 15 & 30 \\\bottomrule
		\footnotesize{$^*$ also for rollout during policy training,} \\
		\footnotesize{$^{**}$ terminates possibly earlier during rollout,} \\
		\footnotesize{$^{***}$ only for subsequent result evaluation}\\
	\end{tabular}
	
	\label{table1}
\end{table}

Table \ref{table1} shows the individual setups for each task. Note that the evaluation horizon possibly differs from the time horizon used during the learning phase. This procedure is useful to assess if the policy is indeed able to keep the target state for an extended period of time. For the \ac{IPSU} task, we use two different horizons. During the first 6 iterations (starting only with the randomly generated initial samples) the planning horizon is reduced to 80 time steps in order to learn the swing up task first. Afterwards, the extended horizon of 110 time steps is used to improve the policy further. All tasks deal with one dimensional action spaces. Hence, the number of trainable parameters of the policy is given as $(\text{number of state dimensions} + 1) \times \text{number of basis functions} + \text{number of state dimensions}$.

\subsection{Computation time} \label{computationtime}

\begin{figure}
	\centering
	\begin{minipage}{0.99\linewidth}
		\centering
		\includegraphics[width=\txtw\textwidth]{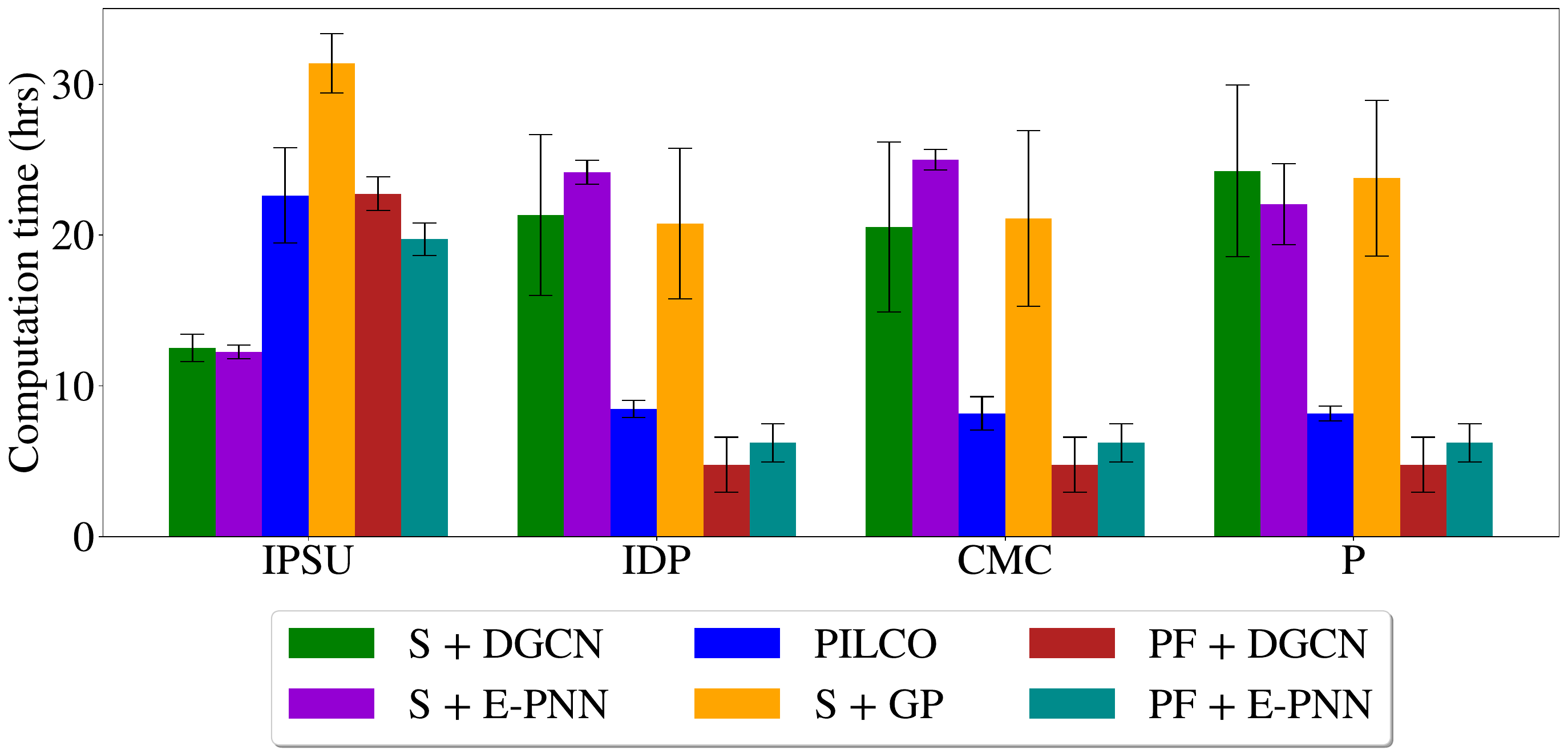}
		\captionof{figure}{Average computation time}\label{fig7}
	\end{minipage}
\end{figure}

Figure \ref{fig7} shows the average computation time and the standard error for all investigated tasks and methods. Please note that the \ac{IPSU} shows very different results than the remaining examples. This fact is caused by multiple reasons. Instead of 15 trials the \ac{IPSU} task used 30 trials and the learning horizon is significantly longer which results in more training data. In particular, the increased amount of data slows down the prediction times of the \ac{GP} and also \ac{PILCO}. Additionally, the policy possesses the highest number of parameters. All these aspects result in a prolongation of the computation time. On the other hand, this particular examples used more computational resources as explained in Section \ref{setup}. Moreover, only 50 instead of 100 trajectories were used for \ac{IPSU} as shown in Table \ref{table1}. For this reason, the trajectory sampling method speeds up and is in this case faster than the density-based uncertainty propagation methods. In \cite{CCAL18} the authors propose using only 20 trajectories, but we decided to use a higher number of particles at the expense of computation time in order to guarantee adequate results. Furthermore, the computation times of some methods have relatively high standard errors which are presumably caused by early stopping of the optimizer.

%
%
%

%

\end{document}